\documentclass[pdflatex,sn-mathphys-num]{sn-jnl}

\usepackage{graphicx}%
\usepackage{multirow}%
\usepackage{amsmath,amssymb,amsfonts}%
\usepackage{amsthm}%
\usepackage{mathrsfs}%
\usepackage[title]{appendix}%
\usepackage{xcolor}%
\usepackage{textcomp}%
\usepackage{manyfoot}%
\usepackage{booktabs}%
\usepackage{algorithm}%
\usepackage{algorithmicx}%
\usepackage{algpseudocode}%
\usepackage{listings}%
\usepackage{subcaption}%

\theoremstyle{thmstyleone}%
\theoremstyle{thmstyletwo}%

\theoremstyle{thmstylethree}%

\begin{document}

\title[Article Title]{Tactical Decision Making for Autonomous Trucks by Deep Reinforcement Learning with Total Cost of Operation Based Reward}


\author*[1,3]{\sur{Deepthi Pathare}}\email{pathare@chalmers.se}

\author[2,3]{\sur{Leo Laine}}\email{leo.laine@chalmers.se}

\author[1]{\sur{Morteza Haghir Chehreghani}}\email{morteza.chehreghani@chalmers.se}

\affil[1]{\orgdiv{Department of Computer Science and Engineering}, \orgname{Chalmers University of Technology and and University of Gothenburg}, \orgaddress{ \city{Göteborg}, \postcode{41296}, \country{Sweden}}}

\affil[2]{\orgdiv{Department of Mechanics and Maritime Sciences}, \orgname{Chalmers University of Technology}, \orgaddress{ \city{Göteborg}, \postcode{41296}, \country{Sweden}}}

\affil[3]{\orgdiv{Safe and Efficient Driving}, \orgname{Volvo Group of Trucks Technology}, \orgaddress{\city{Göteborg}, \postcode{41715}, \country{Sweden}}}

\abstract{We develop a \emph{deep reinforcement learning} framework for tactical decision making in an autonomous truck, specifically for Adaptive Cruise Control (ACC) and lane change maneuvers in a highway scenario. Our results demonstrate that it is beneficial to separate high-level decision-making processes and low-level control actions between the reinforcement learning agent and the low-level controllers based on physical models. In the following, we study optimizing the performance with a realistic and multi-objective reward function based on Total Cost of Operation (TCOP) of the truck using different approaches; by adding weights to reward components, by normalizing the reward components and by using curriculum learning techniques.}




\maketitle

\section{Introduction}\label{sec1}

The efficiency and safety of transport networks have a huge impact on the socio-economic development of the globe. A significant part of this network is freight transport, of which more than $70\%$ is carried out by trucks \cite{de2004national}. The modeling of these complex networks, with a particular focus on traffic scenarios, including trucks and their Total Cost of Operation (TCOP) is thus of paramount importance to developing sustainable traffic solutions.

At a mesoscopic scale, a truck can significantly affect the surrounding traffic \cite{moreno2018stability}, primarily due to its comparatively larger size and length \cite{yang2015cellular}. It also needs more cooperation from surrounding vehicles in order to perform specific maneuvers, such as lane change in multiple-lane dense traffic scenarios \cite{nilsson2017traffic}. Further, if it is a Long Combination Vehicle (LCV), its influence on the safety and evolution of the surrounding traffic is substantial \cite{grislis2010longer}.

Modern vehicles, including trucks, are equipped with a number of features that enhance their performance at different levels. For example, modern trucks operate in connected networks, constantly exchanging data related to their location and performance with their clients. They are also equipped with Advanced Driver Assistance Systems (ADAS) to assist the driver with complex driving tasks and to enhance safety \cite{shaout2011advanced,jimenez2016advanced}. Adaptive Cruise Control (ACC) is such a driver assistance function that provides longitudinal control and maintains a safe distance with the vehicle ahead \cite{xiao2010comprehensive}.

Artificial Intelligence (AI) and Machine Learning have revolutionized the connectivity and autonomy of vehicles, including trucks. With the integration of sensors, cameras, and sophisticated onboard systems, machine learning algorithms can analyze vast amounts of real-time data, allowing trucks to make intelligent decisions on the road. Continuous advancements in machine learning continue to push the boundaries of connectivity and autonomy, transforming the trucking industry towards a more efficient and intelligent future.

A widely used machine learning framework in the context of autonomous systems is Reinforcement Learning (RL). RL overcomes the challenges of traditional search based methods such as A* search, which lack the ability to generalize to unknown situations in a non-deterministic environment and are computationally expensive \cite{Sutton1998}. RL methods also have significant advantages over traditional car-following and lane change models which are based on predefined rules and assumptions. Behavior of traditional models are deterministic and do not adapt beyond their programmed logic. On the otherhand, advanced RL methods such as Deep Reinforcement Learning (DRL) can adapt to non-deterministic environments with varying surrounding vehicle behaviors and make informed decisions in complex scenarios by processing high-dimensional state representations. RL has been increasingly used to solve complex problems related to autonomous car driving, such as navigation, trajectory planning, collision avoidance, and behavior planning \cite{sallab2017deep,shalev2016safe,kiran2021deep}. The application of this framework to autonomous trucks is a relatively new area of research. An important contribution in this direction is the work \cite{hoel2020tactical}, which implements an RL framework for autonomous truck driving in SUMO. They study how a Bayesian RL technique, based on an ensemble of neural networks with additional randomized prior functions (RPF), can be used to estimate the uncertainty of decisions in autonomous driving. On the other hand, more results exist when we consider autonomous driving of passenger cars. The study in \cite{5876320} develops a cooperative adaptive cruise control (CACC) using RL for securing longitudinal following of a front vehicle using vehicle-to-vehicle communication. The CACC system has a coordination layer that is responsible for the selection of high-level actions, e.g., lane changing or secure vehicle following.  The action layer, which is a policy-gradient RL agent, must achieve this action by selecting the appropriate low-level actions that correspond to the vehicle’s steering, brakes, and throttle. Another study in \cite{zhao2013supervised} proposes a Supervised Actor-Critic (SAC) \cite{NIPS1999_6449f44a, haarnoja2018soft} approach for optimal control of ACC. Their framework contains an upper level controller based on SAC, which chooses desired acceleration based on relative velocity and distance parameters of leading and following vehicles. A low level controller receives the acceleration signal and transfers it to the corresponding brake and/or throttle control action. In \cite{9152161}, the authors develop a DRL framework for ACC and compare the results with Model Predictive Control. In all these studies, the RL agent performs step control of acceleration or other driving maneuvers such as braking and steering. For this reason, it can take longer time for the agent to achieve maximum speed even though there is no leading vehicle in front, making the process apparently inefficient.

There is also only limited prior work available where the RL agent is trained to learn high level actions such as choosing the safe gap with the leading vehicle and the actual speed control is performed by a low level controller. The work in \cite{das2021saint} develops a similar system that aims to achieve simultaneous optimization of traffic efficiency, driving safety, and driving comfort through dynamic adaptation of the inter-vehicle gap. It suggests a dual RL agent approach - the first RL agent is designed to find and adapt the optimal time-to-collision (TTC) threshold based on rich traffic information, including both macroscopic and microscopic traffic data obtained from the surrounding environment and the second RL agent is designed to derive an optimal inter-vehicle gap that maximizes traffic flow, driving safety, and comfort at the same time. In \cite{9338516}, authors develop a Deep Deterministic Policy Gradient (DDPG) - Proportional Integral Derivative (PID) controller for longitudinal control of vehicle platooning. They use the DDPG algorithm to optimize the $K_{p}$, $K_{d}$ and $K_{i}$ constants for the PID controller which controls the desired speed. Both these studies mainly focus on longitudinal control of vehicles, and the effects of lane change maneuvers are not studied.

We develop ACC, together with lane change maneuvers for an autonomous truck in a highway scenario. We propose an architecture that combines RL with low-level controllers and separate the high level and low level decision making between them to improve safety and efficiency. The results of this architecture are compared to a baseline architecture that solely relies on RL to perform actions. We evaluate the performance with three different RL algorithms: Deep Q-Network (DQN) \cite{Mnih2015HumanlevelCT}, Advantage Actor-Critic (A2C) \cite{pmlr-v48-mniha16} and Proximal Policy Optimization (PPO) \cite{schulman2017proximal}. We design two reward functions for training the RL agents. The first focuses on safety, while the second is based on Total Cost of Operation (TCOP), covering realistic expenses like energy consumption and human resources. We investigate different training methods to handle the complex and realistic reward function based on TCOP. This approach establishes a strong foundation for future research into improving the economic viability and operational efficiency of autonomous driving systems.

This work is an extension of our previous work \cite{10386803} wherein we additionally, (i) investigate training of RL agent with the complex TCOP based reward function consisting of realistic costs and revenue values with and without normalization (ii) extend our RL framework by incorporating curriculum learning techniques \cite{bengio2009curriculum, narvekar2020curriculum} and compare the results with the non-curriculum learning approach. All our results are obtained using Simulation of Urban Mobility (SUMO) \cite{SUMO2018}, which is a widely used simulation platform for the conceptual development of autonomous vehicles. The tailored RL environment based on SUMO that we developed for heavy vehicle combination driving in highway traffic is provided as open access, which offers a great open source framework for investigating various RL methods in complex settings.\footnote{Source Code: \url{https://github.com/deepthi-pathare/Autonomous-truck-sumo-gym-env}}

\section{Tactical Decision Making with Reinforcement Learning}\label{method}
 
\subsection{Environment Setup}
\begin{figure}[htb!]
      \centering
      \includegraphics[scale=0.47, trim={0, 0, 0, 0}, clip] {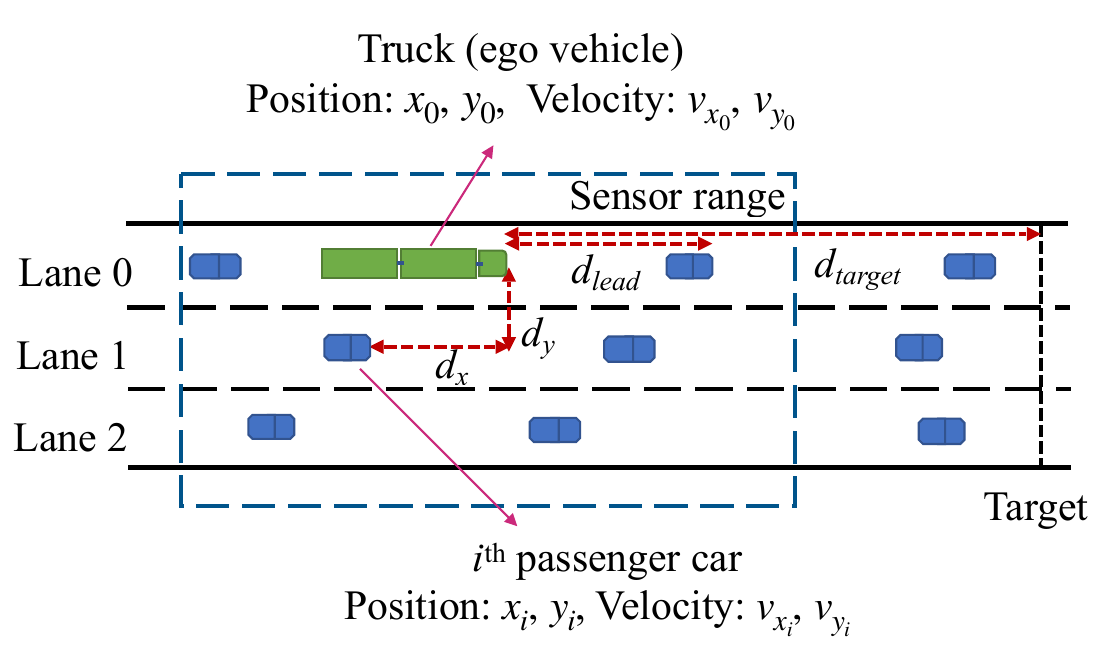}
      \caption{Schematic diagram of the highway simulation environment. The truck in green color is the ego vehicle.}
      \label{fig:env}
    \vspace{-5mm}    
\end{figure}

In this work, we consider a dynamic highway traffic environment with an autonomous truck (ego vehicle) and passenger cars as shown in Fig. \ref{fig:env}. The maximum speed of the ego vehicle is set to be 25 m/s. Furthermore, 15 passenger cars with speed between 15 m/s and 35 m/s are simulated based on the default Krauss car following model \cite{Krauss1997} and LC2013 lane change model \cite{dlr89233} in SUMO. The initial position and speed pattern of the surrounding vehicles are set randomly which makes the surrounding traffic nondeterministic throughout the simulation. Starting from the initial position (which will be 800m after initialization steps in SUMO), the ego vehicle is expected to reach the target set at a distance of 3000m. This means that the ego vehicle has to drive 2200m on the highway in each episode. An episode will also terminate if a hazardous situation such as a collision or driving outside the road happens or if a maximum of 500 RL steps are executed. The observation or state of the environment at each step contains information about position, speed, and state of left/right indicators for ego vehicle and vehicles within the sensor range. The sensor range used here is 200m.

\subsection{Reinforcement Learning Framework}
Reinforcement learning is a branch of machine learning where an agent acts in an environment and tries to learn a policy, $\pi$, that maximizes a cumulative reward function. The policy defines which action, $a$, to take, given an observation or state, $s$. This action leads to a new state of the environment $s^\prime$, and returns a reward, $r$.

\subsubsection{MDP Formulation}
A reinforcement learning problem can be modeled as a Markov Decision Process (MDP), which is defined by the tuple (S, A, T, R, $\gamma$), where S is the state space, A is the action space, P is the transition dynamics, R is the reward model, and $\gamma$ is a discount factor. The tactical decision making problem for autonomous truck in this study is formulated as MDP as follows:
\begin{itemize}
\item \textbf{State Space (S):} The state includes observations from ego vehicle and the surrounding vehicles.
Following are the observations for the ego vehicle:
\begin{enumerate}
    \item Longitudinal speed ($v_{x0}$)
    \item Lane change state ($sign(v_{y0})$)
    \item Lane number
    \item State of left indicator 
    \item State of right indicator
    \item Target(leading) vehicle distance ($d_{lead}$)
\end{enumerate}

Following are the observations for each vehicle in the sensor range of the ego vehicle:
\begin{enumerate}
    \item Relative longitudinal distance from ego vehicle ($d_{xi}$)
    \item Relative lateral distance from ego vehicle ($d_{yi}$)
    \item Relative longitudinal speed with ego vehicle ($v_{xi} - v_{x0}$)
    \item Lane change state ($sign(v_{yi})$)
    \item Lane number
    \item State of left indicator
    \item State of right indicator
\end{enumerate}

\item \textbf{Action Space (A):} The action space is defined separately for the baseline and new architectures described in section \ref{sec:arch}.
\\
\item \textbf{Transition Dynamics(P):} The transition dynamics is defined by SUMO, and is not known to the RL agent.
\\
\item \textbf{Reward Function(R):} 
We design a basic reward function focused on safety aspects to motivate the agent to drive at maximum speed and reach the target without any hazardous situations. This reward function is similar to that in \cite{hoel2020tactical} except that a reward for reaching the target is added, whereas the penalty for emergency braking is not considered for simplicity. The reward at each time step is given by (\ref{eq:base_rew}).
\begin{equation}
\begin{aligned}
\label{eq:base_rew}
    r(t) = \frac{v_{t}}{max\_v} - I_l\:P_l-I_c\:P_c-I_{nc}\:P_{nc}-I_o\:P_o\\+I_{tar}\:\frac{R_{tar}}{T}   
\end{aligned}
\end{equation}
Here $v_{t}$ is the velocity of the vehicle at time step $t$ and $max\_v$ is the maximum velocity of the vehicle. $I$ is an indicator function, which takes a value of 1 when the corresponding condition is satisfied and $P$ and $R$ are the corresponding penalty and reward values respectively. The possible conditions are lane change ($l$), collision ($c$), near collision ($nc$), driving outside the road ($o$) and reaching the target ($tar$). $T$ is the total time it takes to reach the target. The parameter values used are given in \autoref{tab:basic_rew_params}.

\begin{table}[htb!]
\caption{Parameter values used in the basic reward function.}
\label{tab:basic_rew_params}
\begin{tabular}{|p{3cm} | p{3cm}|}
\hline
\textbf{Parameter} & \textbf{Value}\\
\hline
$P_l$&1
\\ 
\hline
$P_c$&10
\\ 
\hline
$P_{nc}$&10
\\ 
\hline
$P_o$&10
\\ 
\hline
$R_{tar}$&100
\\ 
\hline
\end{tabular}
\end{table}

Furthermore, we have also designed a complex and multi-objective reward function based on TCOP which is described in section \ref{sec:tcop_rew}.
\\
\item \textbf{Discount Factor($\gamma$):} $\gamma$ is set to 0.99.
\end{itemize}

\subsubsection{Reinforcement Learning Algorithms}

We choose three reinforcement learning algorithms as listed below for the decision making in baseline and new architectures. The implementations of these algorithms from the stable-baselines3 library with the default hyperparameters are used \cite{stable-baselines3}.

\begin{enumerate}
    \item \textbf{Deep Q-Network (DQN)} \\
DQN is a reinforcement learning algorithm based on the Q-learning algorithm, which learns the optimal action-value function by iteratively updating estimates of the Q-value for each state-action pair. In DQN,  the optimal action-value function in a given environment is approximated using a neural network model. The corresponding objective function is given by,
\begin{align}
    L(\theta) = \mathbb{E}_{s,a,r,s'}\Big[\Big(y - Q(s,a;\theta)\Big)^2\Big]
\end{align}

where $\theta$ represents the weights of the Q-network, $s$ and $a$ are the state and action at time $t$, $r$ is the immediate reward, $s'$ is the next state, and $y = r + \gamma \max_{a'} Q(s',a';\theta^-)$ is the target value. The target value is the sum of the immediate reward and the discounted maximum Q-value of the next state, where $\gamma$ is the discount factor and $\theta^-$ represents the weights of a target network with delayed updates. The objective function is to minimize the mean squared error between the estimated Q-value and the target value, which is then optimized using stochastic gradient descent.

Another improvement of DQN over the standard Q learning approaches is the use of a replay buffer and target network. The agent uses replay buffer to store transitions and samples from it randomly during training. This enables the efficient use of past experiences. The target network is a copy of the Q-network with delayed updates. Together with the replay buffer, it improves the learning stability of the model and help prevent over fitting.

\item \textbf{Advantage Actor-Critic (A2C)} \\
A2C is a reinforcement learning algorithm that combines the actor-critic method with an estimate of the advantage function. In the actor-critic method, the agent learns two neural networks: an actor network that outputs a probability distribution over actions, and a critic network that estimates the state-value function. The actor network is trained to maximize the expected reward by adjusting the policy parameters, while the critic network is trained to minimize the difference between the estimated value function and the true value function. Advantage function calculates how better taking an action at a state is compared to the average value of the state which is computed as below:

\begin{equation}
A(s_t,a_t) = r_{t+1} + \gamma V(s_{t+1}) - V(s)
\end{equation}

Actor uses the computed advantage function value from critic as a feedback and updates the policy parameters $\theta$ as,

\begin{equation}
\Delta \theta=\alpha_\theta A(s_{t}, a_{t}) \nabla_\theta\left(\log \pi_\theta(s_t, a_t)\right) 
\end{equation}

Critic updates its value function parameters $w$ as,

\begin{equation}
\Delta w=\alpha_w A(s_t, a_t )\nabla_w \hat{v}_w\left(s_t\right)
\end{equation}

Here, $\alpha_\theta$ and $\alpha_w$ are the learning rates.

\item \textbf{Proximal Policy Optimization (PPO)} \\
 PPO belongs to the family of policy gradient algorithms and has exhibited great potential in effectively addressing diverse RL problems \cite{schulman2017proximal,svensson2023utilizing}. In addition, PPO benefits from simple yet effective implementation and a broad scope of applicability. 

The PPO algorithm is designed to address the challenges such as the instability that can arise when the policies are rapidly updated. PPO works by optimizing a surrogate objective function that measures the difference between the updated policy and the previous one. 
The surrogate objective function of PPO is defined as the minimum of two terms: the ratio of the new policy to the old policy multiplied by the advantage function, and the clipped ratio of the new policy to the old policy multiplied by the advantage function. In mathematical terms, this function is given as, 
\begin{equation}
\begin{aligned}
L^{CLIP}(\theta) = \hat{{E}}_t\Big[\min\Big(r_t(\theta)\hat{A}_t, \text{clip}(r_t(\theta), 1-\epsilon, 1+\epsilon)\hat{A}_t\Big)\Big]
\end{aligned}    
\end{equation}
where $\theta$ represents the policy parameters, ${r_t(\theta) = \frac{\pi_\theta(a_t|s_t)}{\pi_{\theta_{old}}(a_t|s_t)}}$ is the likelihood ratio between the current policy and the previous policy, $\hat{A}_t$ is the estimated advantage of taking action $a_t$ in state $s_t$ and $\epsilon$ is a hyperparameter that controls the degree of clipping. The clipping term ensures that the policy update does not result in significant policy changes, which can destabilize the learning process. The PPO algorithm maximizes this objective function using stochastic gradient ascent to update the policy parameters. The expectation is taken over a batch of trajectories sampled from the environment.
\end{enumerate}

\subsection{System Architectures} \label{sec:arch}
\subsubsection{Baseline architecture}
\begin{figure}
\centering
  \centering
  \includegraphics[scale=0.43]{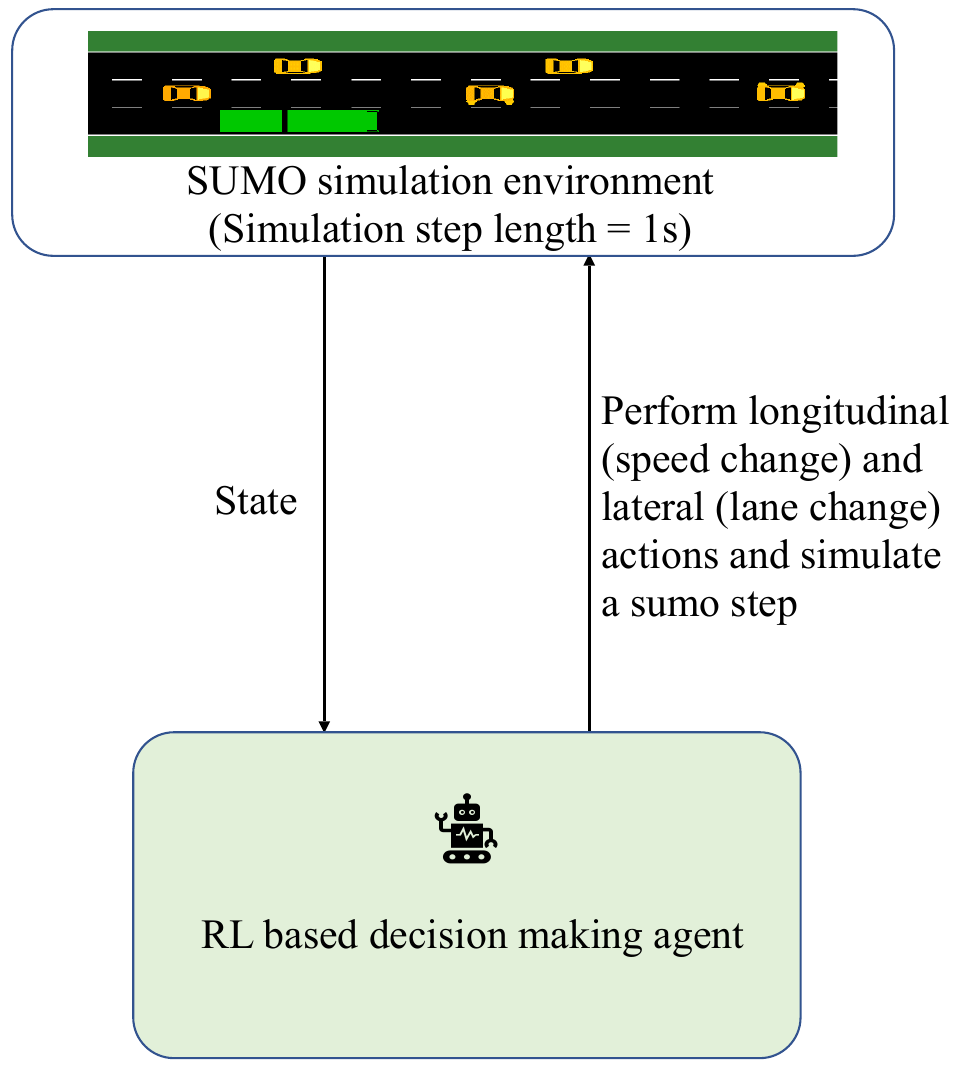}
  \caption{Overview of the baseline architecture}
  \label{fig:sys_design_base}
\end{figure}

The baseline architecture we consider in this work is inspired from \cite{hoel2020tactical} and is illustrated in Fig. \ref{fig:sys_design_base}. Here the decision making is done solely by an RL agent, every 1s. The action space for the agent is discrete. Each action consists of a combination of longitudinal action and lateral action. Longitudinal actions are speed changes of 0, 1, -1 or -4 m/s. Lateral actions are to stay on lane, change to left lane, or change to right lane. In this paper, we compare the performance of this baseline architecture with our new architecture described below.

\subsubsection{New architecture}

\begin{figure}
  \centering
  \includegraphics[scale=0.4]{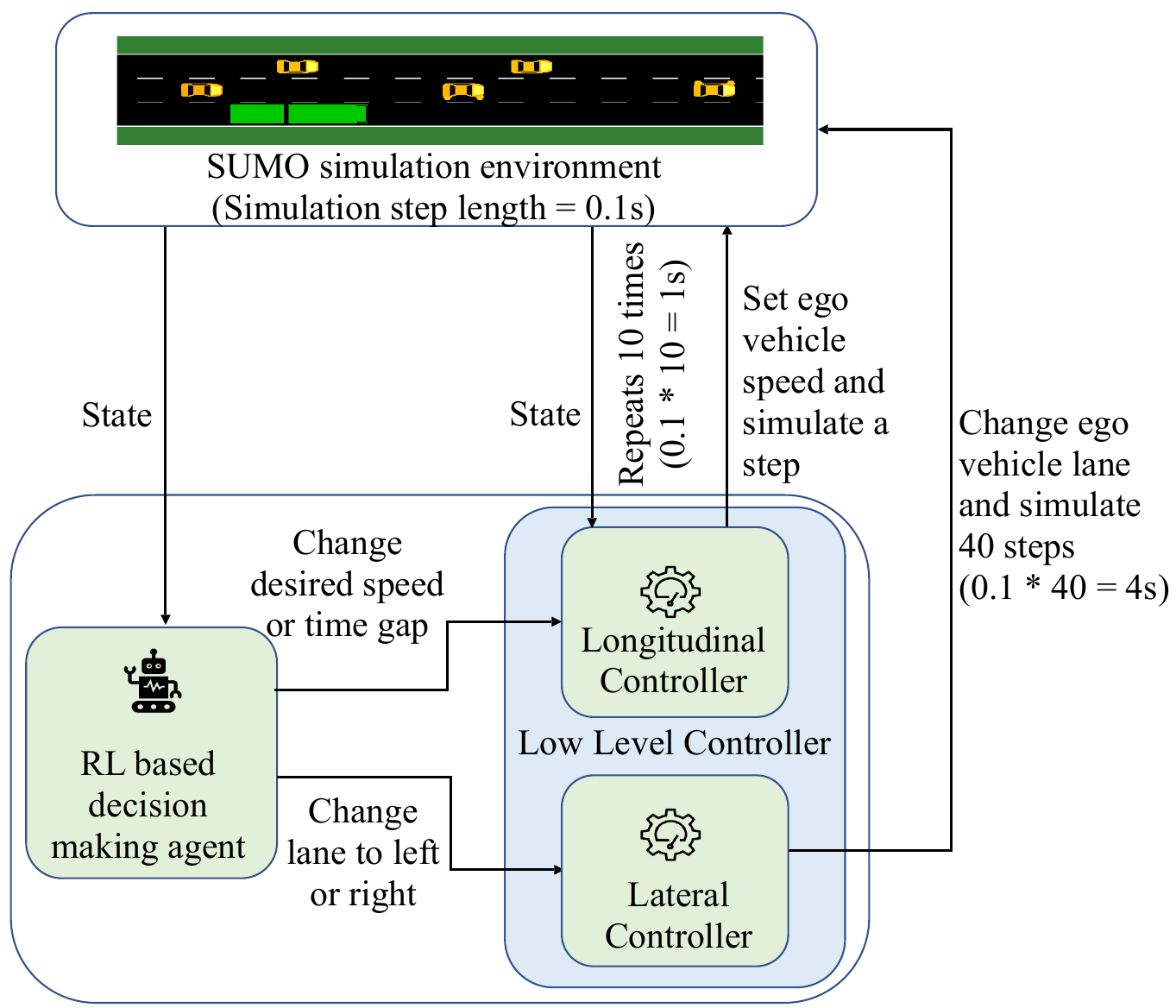}
  \caption{Overview of the new architecture}
  \label{fig:sys_design}
\end{figure}

Fig. \ref{fig:sys_design} shows the new architecture diagram for the automated truck driving framework. It has mainly three components; a high level decision making agent based on RL, a low level controller for longitudinal control and a low level controller for lateral control. The RL agent chooses a longitudinal action or a lateral action based on the current state of the SUMO environment. 

The action space is defined below.
\begin{enumerate}
    \item Set short time gap with leading vehicle (1s)
    \item Set medium time gap with leading vehicle (2s)
    \item Set long time gap with leading vehicle (3s)
    \item Increase desired speed by 1 m/s
    \item Decrease desired speed by 1 m/s
    \item Maintain current desired speed and time gap
    \item Change lane to left
    \item Change lane to right
\end{enumerate}

The longitudinal action can be one of the following: set desired time gap (short, medium, long), increment or decrement desired speed or maintain the current time gap and desired speed. When RL agent chooses one of these actions, the longitudinal controller will compute acceleration of ego vehicle based on Intelligent Driver Model (IDM) \cite{treiber2000congested} given by,

\begin{equation}
\begin{aligned}
& \dot{v}_\alpha=\frac{\mathrm{d} v_\alpha}{\mathrm{d} t}=a\left(1-\left(\frac{v_\alpha}{v_0}\right)^\delta-\left(\frac{s^*\left(v_\alpha, \Delta v_\alpha\right)}{s_\alpha}\right)^2\right), \\
& s^*\left(v_\alpha, \Delta v_\alpha\right)=s_0+v_\alpha T+\frac{v_\alpha \Delta v_\alpha}{2 \sqrt{a b}}
\label{eq:idm}
\end{aligned}
\end{equation}
\noindent
where $\alpha$ is the ego vehicle and $\alpha - 1$ is the leading vehicle. $v$ denotes the velocity and $l$ denotes the length of the vehicle. $s_\alpha:=x_{\alpha-1}-x_\alpha-l_{\alpha-1}$ is the net distance and $\Delta v_\alpha:=v_\alpha-v_{\alpha-1}$ is the velocity difference. $v_0$ (desired velocity), $s_0$ (minimum spacing), $T$ (desired time gap), $a$ (maximum acceleration), and $b$ (comfortable braking deceleration) are model parameters. 

 Here IDM uses the latest desired speed and time gap set by the RL agent. It computes a new acceleration for the truck and sets the resulting speed in SUMO every 0.1s. This process continues for a total duration of 1s, after which the RL agent chooses the next high level action. Note that in the new architecture, the ACC mode is always activated for the truck by construction.

The lateral action is to change the truck's course to left or right lane. When the RL agent chooses a lateral action, the lateral controller initiates the lane change. Lane change is performed using the default LC2013 lane change model \cite{dlr89233} in SUMO. The lane width is set to $3.2m$ and the lateral speed of the truck is set to $0.8 m/s$. Hence, in total, it takes $4s$ to complete a lane change action, which corresponds to 40 SUMO time steps of $0.1s$ duration. Following this, the RL agent chooses the next high level action.

\subsection{Total Cost of Operation}\label{sec:tcop_rew}
The TCOP of a truck encompasses various expenses incurred during its operation such as energy cost, driver cost and insurance cost. We have designed a TCOP-centric reward function to provide the agent with more realistic goals similar to those of a human truck driver and to motivate the agent to learn safe and cost-effective actions.  The reward at each time step is given by (\ref{eq:tcop_rew}). Here, we consider realistic cost or revenue values (in euros) for each component to see if the agent can learn a safe and cost effective driving strategy with this reward function.

\begin{equation}
\begin{aligned}
\label{eq:tcop_rew}
r(t) = - C_{el}\:e_t - C_{dr}\:\Delta t - I_l\:P_l-I_c\:P_c\:W_c \\ -I_{nc}\:P_{nc}\:W_{nc}-I_o\:P_o\:W_o + I_{tar}\:R_{tar}\:W_{tar}
\end{aligned}
\end{equation}
$C_{el}$ is the electricity cost, $e_t$ is the electricity consumed at time step $t$, $C_{dr}$ is the driver cost and $\Delta t$ is the duration of a time step. The electricity consumed during the time step $t$ ($e_t$) is calculated as,
\begin{equation}
\begin{aligned}
\label{eq:energy}
    e_t = f_t\:v_t\:\Delta{t},
\end{aligned}
\end{equation}
where $f_t$, force at time step $t$ is given by, 
\begin{equation}
\begin{aligned}
f_t = m\:a_t + \frac{1}{2} C_d\:A_f \: \rho_{\text{air}} \: v^2 + m \: g \: C_r \\ +m \: g \: \sin (\arctan ( \frac{\text{slope}}{100}))
\end{aligned}
\end{equation}
Here $m$ is the mass of the vehicle, $C_d$ is the coefficient of air drag, $A_f$ is the frontal area, $\rho_{\text{air}}$ is the air density, $C_r$ is the coefficient of rolling resistance, $g$ is the acceleration due to gravity and $a$ is the acceleration of the vehicle at time step $t$. The slope of the road is set to be 0 in this study. The parameter values used are given in \autoref{tab:rew_params}.

\begin{table}[htb!]
\caption{Parameter values used in TCOP based reward function.}
\label{tab:rew_params}
\begin{tabular}{|p{2.2cm} | p{2.2cm}|p{2.2cm}|}
\hline
\textbf{Parameter} & \textbf{Value}\\
\hline
$P_l$&0.1
\\ 
\hline
$P_c$&1000 euros
\\ 
\hline
$P_{nc}$&1000 euros
\\ 
\hline
$P_o$&1000 euros
\\ 
 \hline
$R_{tar}$&2.78 euros
\\ 
\hline
$C_{el}$&0.5 euro per kwh
\\ 
\hline
$C_{dr}$&50 euro per hour
\\ 
\hline
$\Delta t$& 1 s
\\ 
\hline
$m$&40000 kg
\\ 
\hline
$C_d$&0.36
\\ 
\hline
$A_f$& 10 $m^2$
\\ 
\hline
$\rho_{\text{air}}$&1.225 $kg/m^3$
\\ 
\hline
$g$& 9.81 $m/s^2$
\\ 
\hline
$C_r$& 0.005
\\ 
\hline
\end{tabular}
\vspace{-2mm}
\end{table}

In the reward function (\ref{eq:tcop_rew}), the penalties are defined as the average cost incurred during hazardous situations. The average cost considered here is the \textit{own risk payment} which is the portion of an insurance claim made by the vehicle owner or the insured for any loss and or damage that occurs when submitting a claim. Similarly, the reward $R_{tar}$ denotes the revenue that can be achieved by the truck when it completes the target. Revenue is computed as the total expected cost with $20\%$ profit in an ideal scenario where the truck drives with an average speed of $22 m/s$ and zero acceleration. The total length of travel is 2200 meters and hence the time to reach the target will be $100s$. Based on (\ref{eq:energy}) and parameter values from Table \ref{tab:rew_params}, the total energy consumed can be computed as $1.85$ kwh. Then, the total energy cost will be $1.85$ kwh $\times 0.5 = 0.925$ euros. The total driver cost for $100s$ will be $1.39$ euros and the total cost becomes $0.925 + 1.39 = 2.315$ euros. Further, adding the $20\%$ profit gives the net revenue $ 2.78$ euros, which is used in the reward function along with a weight $W_{tar}$. We also added weight components $W_c$, $W_{nc}$, $W_o$ for the penalties for collision, near collision and driving outside the road condition respectively.

\subsection{Curriculum Reinforcement Learning}
Curriculum Reinforcement Learning (CRL) is a training strategy where the model is gradually exposed to increasingly complex tasks or difficult examples over time. This approach is based on the intuition that starting with simpler tasks allows the model to build foundational knowledge and gradually progress to more challenging tasks, which can help improve learning efficiency. 

Effectiveness of curriculum learning has been investigated in a number of problems with complex tasks related to autonomous driving, though different from our setting. The study in \cite{9782734} introduces multi-stage learning in complex driving environments. Different stages include a diverse set of starting locations, varying weather conditions, and dense traffic scenarios. The action space consists of accelerator or brake values and steering angles. The reward function penalizes collisions, following the wrong route, and exceeding speed limits, guiding the agents towards safe and efficient driving behaviors. The papers \cite{9789217} and \cite{proc9403} utilize a curriculum learning approach for overtaking in autonomous driving. \cite{9789217} employs a two-stage successive learning progression, where agents first learn to drive as fast as possible and then master the skill of overtaking efficiently. On the other hand, \cite{proc9403} introduces a three-stage curriculum learning process where agents first learn to race, then focus on overtaking maneuvers, and finally refine their skills to race at high speeds while avoiding collisions. Both of these works demonstrate that curriculum learning boosts convergence and contributes to a better final policy. 

The above mentioned studies do not address the problem of strategic decision making for optimizing cost, efficiency and safety at the same time. We address this by combining curriculum learning with RL in the new system architecture and investigate its effectiveness in training the RL agent with the complex reward function based on TCOP. Here, the curriculum for training the RL agent is designed as follows:

\begin{itemize}
    \item \textbf{Curriculum-1}: Learning longitudinal and lateral control without collision/driving outside the road and by reducing driver cost
    \item \textbf{Curriculum-2}: Learn to minimize energy cost
    \item \textbf{Curriculum-3}: Learn to reach the destination successfully within maximum steps
\end{itemize}

In this CRL approach, we update the reward function in each curriculum whereas the RL environment, state and action space remain the same. In this work, we have redefined TCOP based reward function as shown in \autoref{eq:tcop_rew_new} by removing the penalty for lane change and the weight components such that the reward function only contains actual cost and revenue values during a truck's operation. Training an agent with realistic reward functions can lead to policies that are more aligned with human expectations.

\begin{equation}
\begin{aligned}
\label{eq:tcop_rew_new}
r(t) = - C_{el}\:e_t - C_{dr}\:\Delta t -I_{c}\:P_{c} -I_{nc}\:P_{nc}-I_o\:P_o +I_{tar}\:R_{tar}
\end{aligned}
\end{equation}

Here, the notations and values are the same as in \autoref{eq:tcop_rew}. This reward function is divided into smaller components and added to the reward functions $r_1(t), r_2(t), r_3(t)$ defined in \autoref{eq:cl_rew} for curriculum-1, curriculum-2 and curriculum-3 respectively. 

\begin{equation}
\begin{aligned}
\label{eq:cl_rew}
    r_1(t) &= -I_c\:P_c - I_{nc}\:P_{nc} - I_o\:P_o - C_{dr}\:\Delta t \\
    r_2(t) &= r_1(t) - C_{el}\:e_t \\
    r_3(t) &= r_2(t) +  I_{tar}\:R_{tar}\\
\end{aligned}
\end{equation}

Here, the negative reward components for drive cost and energy cost contradict to each other as driver cost motivates the agent to drive faster while energy cost motivates the agent to drive slower. This will lead to challenges in learning an optimal policy as shown in the experiments section. Therefore, we tuned the reward function for curriculum-1 and curriculum-2 to normalize these two components with the distance travelled per time step $\Delta d$ as shown in \autoref{eq:cl_rew_norm}.

\begin{equation}
\begin{aligned}
\label{eq:cl_rew_norm}
    r_1(t) &= -I_c\:P_c - I_{nc}\:P_{nc} - I_o\:P_o - \frac{(C_{dr}\:\Delta t)}{\Delta d} \\
    r_2(t) &= r_1(t) - \frac{(C_{el}\:e_t)}{\Delta d} \\
\end{aligned}
\end{equation}

\section{Experiments}\label{experiments}

This section presents the results from different experiments conducted on the SUMO platform using the baseline and new architectures with different RL algorithms. 
The experiments are conducted on a linux cluster that consists of 28 CPUs with the model \textit{Intel(R) Xeon(R) CPU E5-2690 v4 @ 2.60GHz}. The average time elapsed for training the new architecture with DQN, A2C and PPO for $1e^6$ timesteps are 12528s (3h 28min), 13662s (3h 47min), 14124s (3h 55min) respectively.

\subsection{Performance improvement based on  selection of states}
\begin{figure}[h!]
 \begin{subfigure}{0.49\textwidth}
     \includegraphics[width=\textwidth]{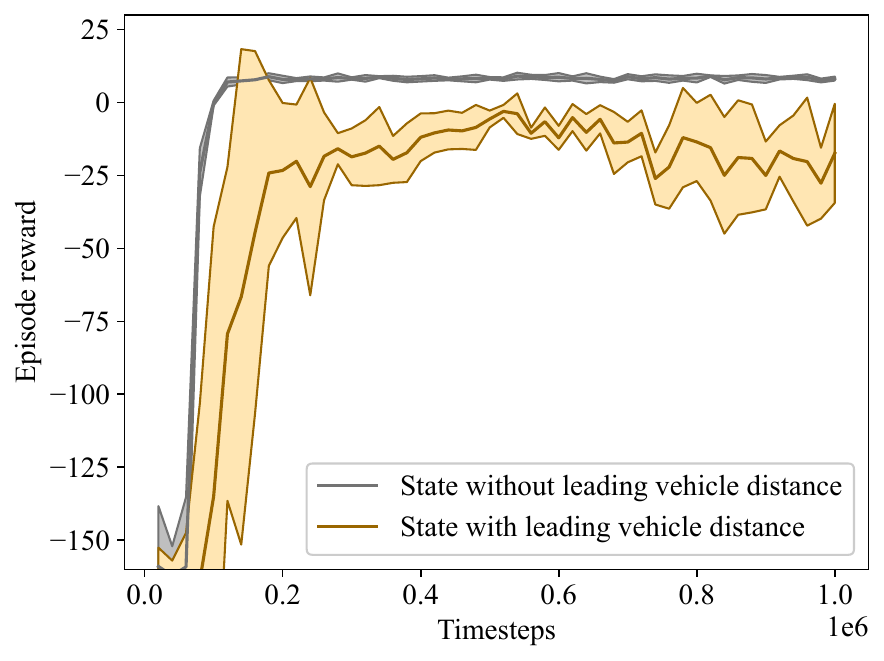}
     \caption{DQN}
     \label{fig:dqn_res}
 \end{subfigure}
 \hfill
 \begin{subfigure}{0.49\textwidth}
     \includegraphics[width=\textwidth]{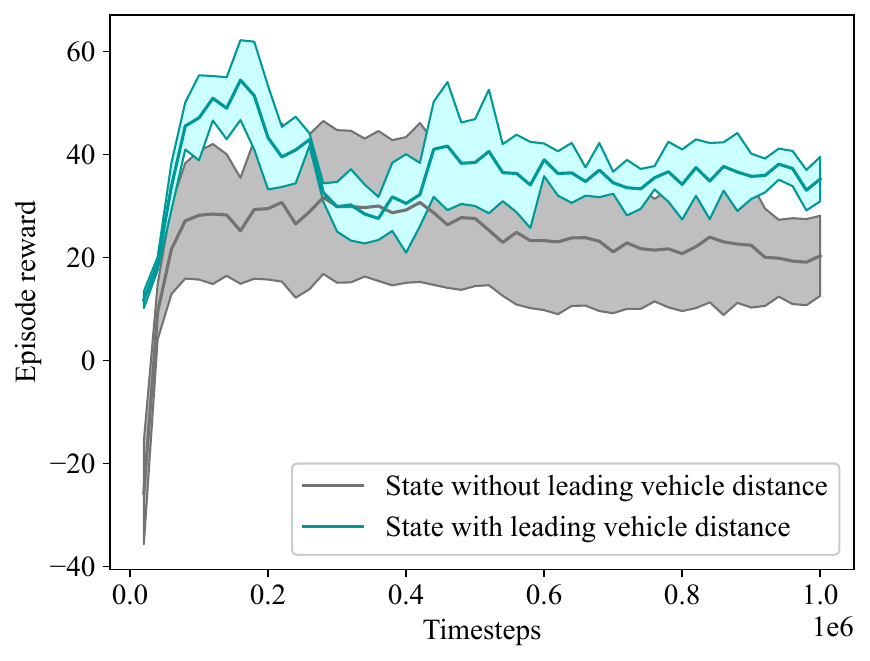}
     \caption{A2C}
     \label{fig:a2c_res}
 \end{subfigure}
 \hfill
 \begin{subfigure}{0.49\textwidth}
     \includegraphics[width=\textwidth]{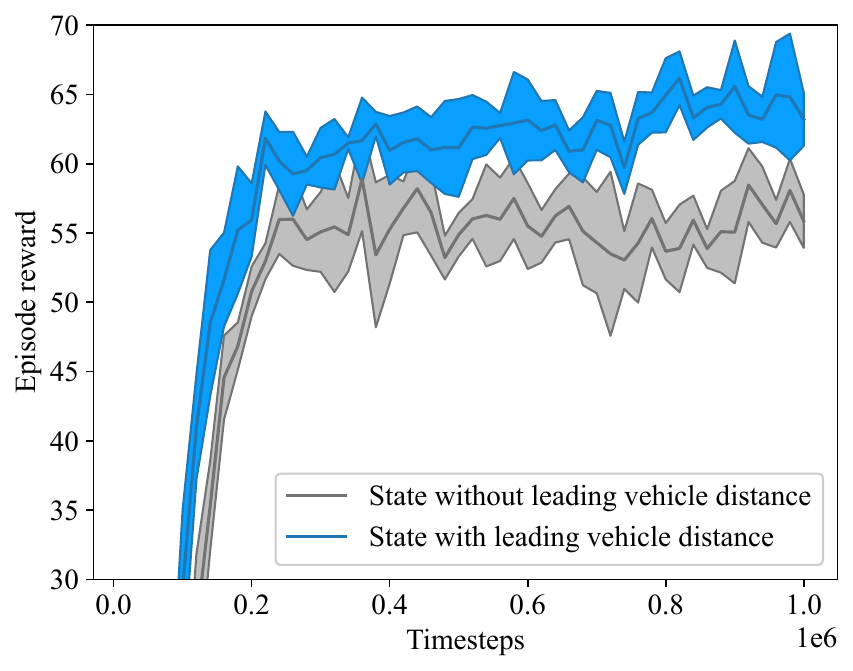}
     \caption{PPO}
     \label{fig:ppo_res}
 \end{subfigure} 
 \caption{Comparison of episode rewards in baseline architecture with and without leading vehicle distance in state space.}
 \label{fig:obs_rew}
\end{figure}

First, we show how the performance can be improved by adding relevant features to the observation space. As mentioned in section II, the state space includes position, speed, and lane change information of the ego vehicle and the surrounding vehicles. In this experiment, we compare the performance of baseline architecture by adding the distance to the leading vehicle as an explicit feature in the above mentioned state space. The basic reward function is used in both cases. The learning curves of average reward (over 5 realizations) using different RL algorithms are shown in Fig. \ref{fig:obs_rew}. The results from PPO and A2C show that the average reward has improved by adding distance to leading vehicle as a feature in the state. Results from DQN show the opposite trend where the average reward for state without leading vehicle distance is higher. However, the average reward values are very low for DQN in both state spaces. and the experiments in upcoming sections also show that DQN is not able achieve good performance for this specific problem.

Table \ref{tab:obs_rew} shows the validation results for PPO algorithm with different evaluation metrics (average of 5 validations with 100 episodes each). The collision rate is considerably reduced and the average speed of the ego vehicle is slightly increased. Note that this distance was already available in the observation space among the properties of surrounding vehicles. Our results show that explicitly adding this as a separate feature helps the agent to learn better how to control the speed w.r.t. the leading vehicle and avoid forward collisions.

\begin{table}[h!]
\caption{Evaluation of baseline architecture using PPO with and without leading vehicle distance in state space.}
\label{tab:obs_rew}
\begin{tabular}{|p{4.9cm} | p{1.5cm}|p{1.5cm}|}
\hline
\textbf{Evaluation Metric} & \textbf{without leading distance} & \textbf{with leading distance}\\
\hline
Reached target successfully & 61\% & 70.6\%
\\ 
\hline
Driven successfully, but not reached target within maximum steps & 0\% & 0\%
\\ 
\hline
Terminated by collision or driving outside road & 39\% & 29.4\%
\\ 
\hline
Average speed & 18.17 m/s & 19.43 m/s
\\ 
\hline
Average distance travelled & 1503.93 m & 1667.86 m
\\ 
\hline
Average executed steps & 88.43 & 89.65
\\ 
\hline
\end{tabular}
\end{table}

\subsection{Performance comparison of the two architectures}
\begin{figure}[h!]
 \begin{subfigure}{0.49\textwidth}
     \includegraphics[width=\textwidth]{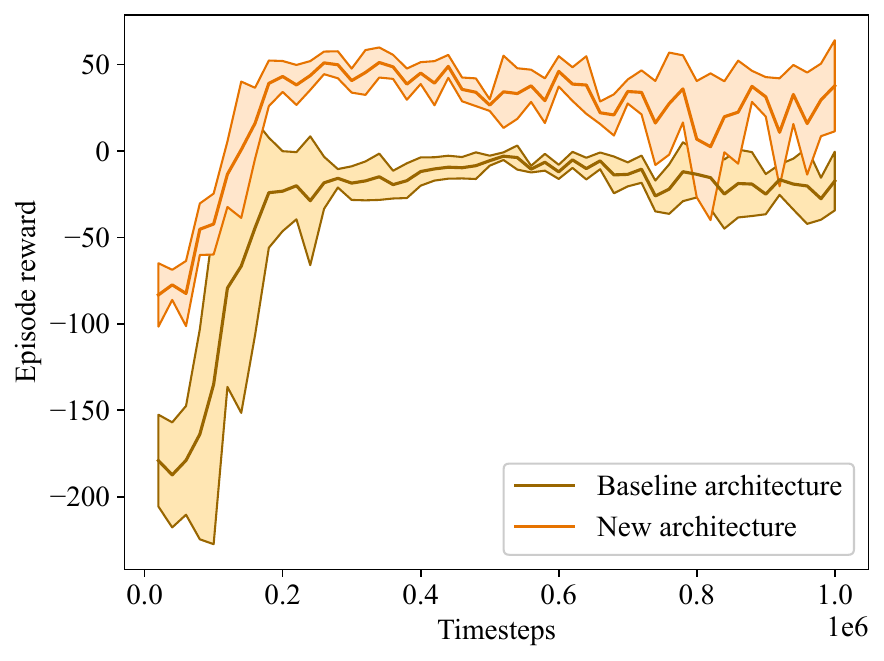}
     \caption{DQN}
     \label{fig:dqn_res}
 \end{subfigure}
 \hfill
 \begin{subfigure}{0.49\textwidth}
     \includegraphics[width=\textwidth]{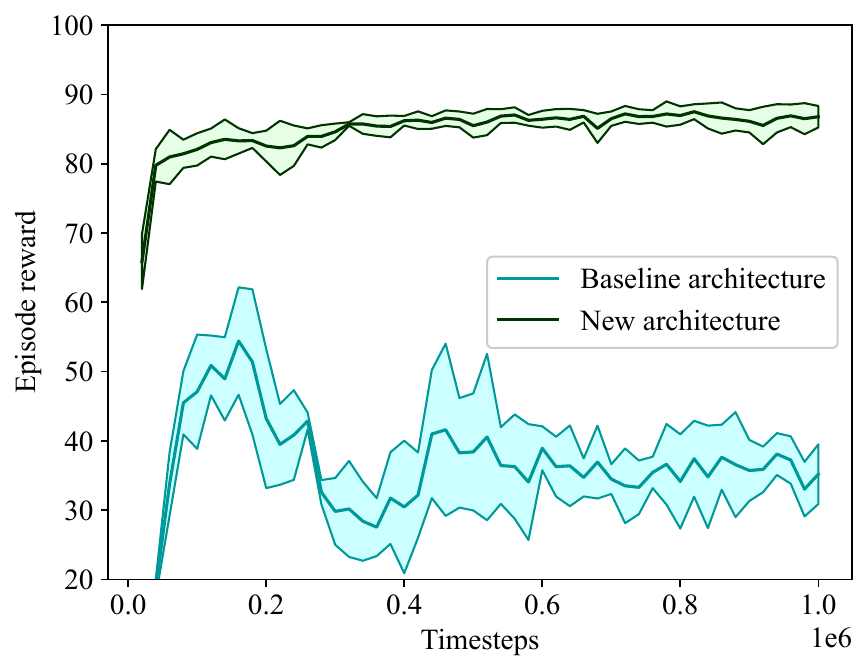}
     \caption{A2C}
     \label{fig:a2c_res}
 \end{subfigure}
 \hfill
 \begin{subfigure}{0.49\textwidth}
     \includegraphics[width=\textwidth]{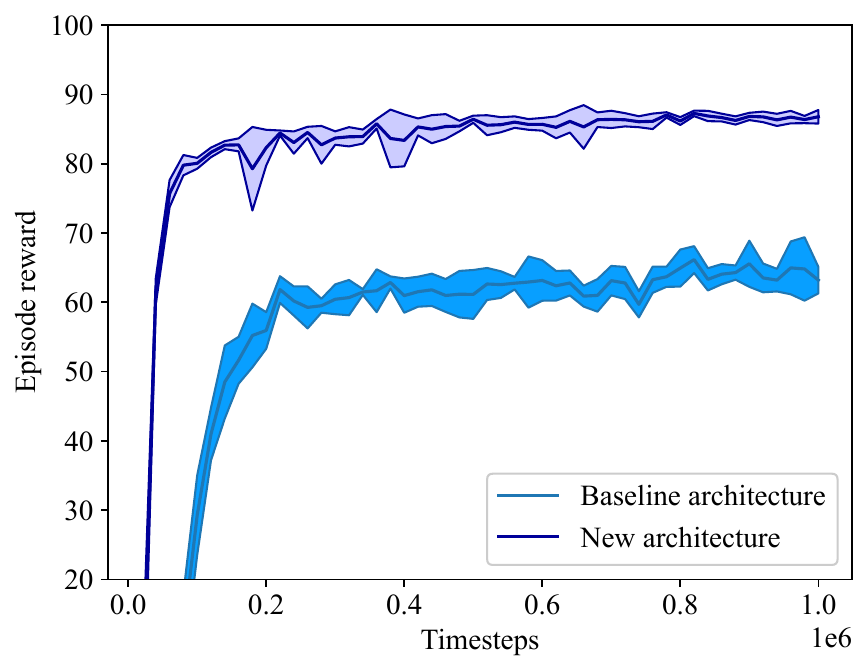}
     \caption{PPO}
     \label{fig:ppo_res}
 \end{subfigure} 
 \caption{Comparison of average episodic reward in the baseline and new architectures with different RL agents}
 \label{fig:arch_comp}
\end{figure}

\begin{figure}[h!]
\begin{subfigure}{0.49\textwidth}
     \includegraphics[width=\textwidth, trim={0, 0, 0, 0}, clip]{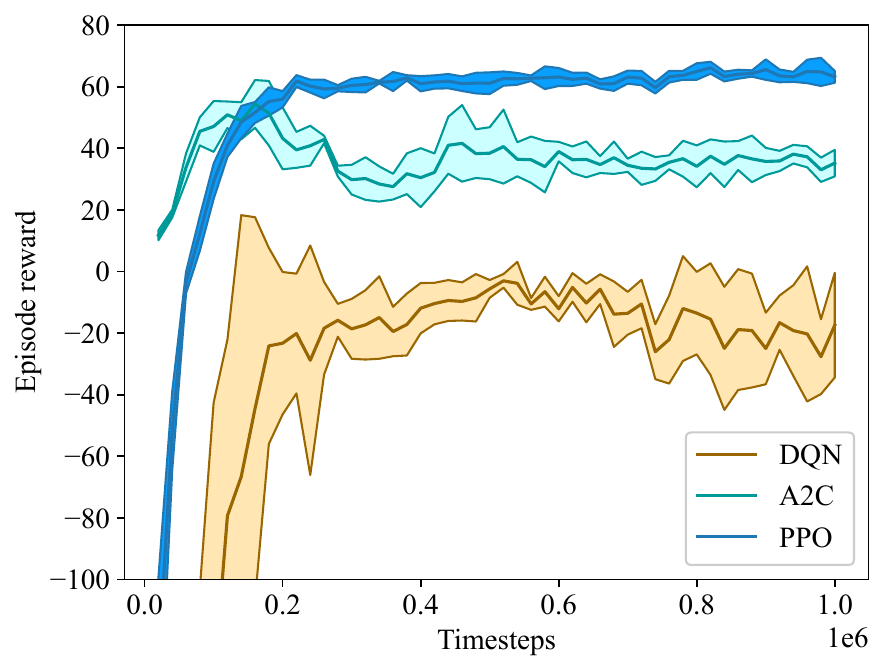}
     \caption{Baseline architecture}
     \label{fig:base_arch_res}
 \end{subfigure}
 \hfill
 \begin{subfigure}{0.49\textwidth}
     \includegraphics[width=\textwidth]{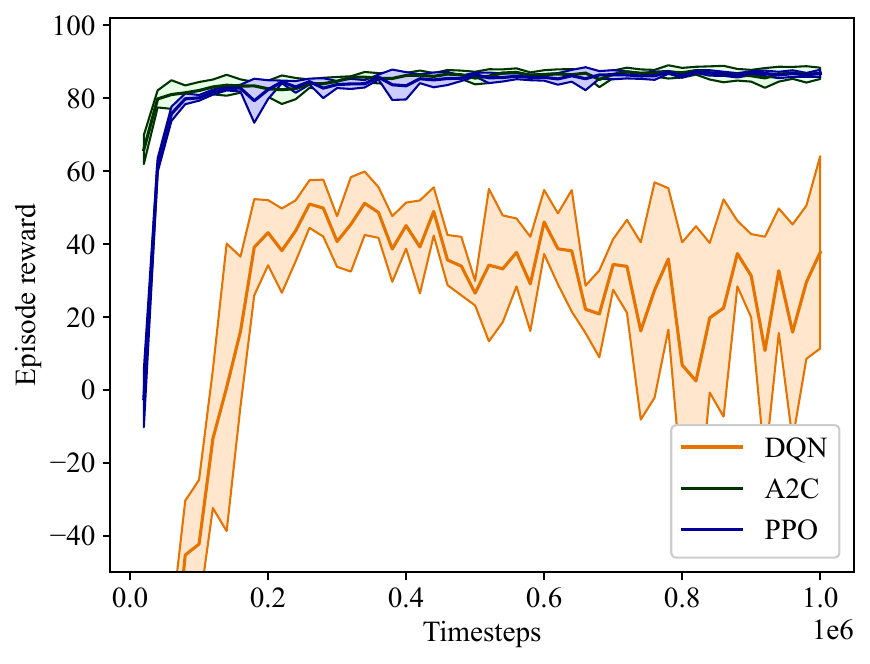}
     \caption{New architecture}
     \label{fig:new_arch_res}
 \end{subfigure}
 \caption{Performance comparison of different RL agents}
 \label{fig:rl_comp}
\end{figure}

In this experiment, we compare the performance of our new architecture shown in Fig. \ref{fig:sys_design} with the baseline framework shown in Fig. \ref{fig:sys_design_base} using different RL algorithms. For both architectures, we use the observation space including leading vehicle distance and the basic reward function. However, the new architecture introduces a low level controller based on a physical model to perform speed control actions. Fig. \ref{fig:arch_comp} shows the comparison of average episode rewards in both architectures for different RL algorithms. It can be seen that the new architecture outperforms the baseline regardless of the chosen RL algorithm. Further, Fig. \ref{fig:rl_comp} compares performance of different RL agents in each architecture. For both cases, DQN obtains lowest average rewards. PPO obtains highest average reward in baseline architecture whereas both PPO and A2C shows similar performance in new architecture.

As PPO gives consistent performance in both architectures, it is used for further validation.The validation results in Table \ref{tab:arch_rew} clearly show that the new architecture has improved the performance compared to the baseline. The collision rate is reduced immensely from 29.4\% to 1.6\%. The slight decrease in the average speed of the new framework is negligible when considering the improvement in safety. The reduced collision rate also explains why there is an increment in the average travelled distance and average executed steps.

\begin{table}[h!]
\caption{Evaluation of baseline and new architectures using PPO.}
\label{tab:arch_rew}
\begin{tabular}{|p{4.9cm} | p{2cm}|p{2cm}|}
\hline
\textbf{Evaluation Metric} & \textbf{Baseline architecture} & \textbf{New architecture}\\
\hline
Reached target successfully & 70.6\% & 97.8\%
\\ 
\hline
Driven successfully, but not reached the target within maximum steps & 0\% & 0.6\%
\\ 
\hline
Terminated by collision or driving outside the road & 29.4\% & 1.6\%
\\ 
\hline
Average speed & 19.43 m/s & 18.56 m/s
\\ 
\hline
Average distance travelled & 1667.86 m & 2178.37 m
\\ 
\hline
Average executed steps & 89.65 & 128.75
\\ 
\hline
\end{tabular}
\end{table}

\subsection{Performance with TCOP based reward}
\subsubsection{Comparison with different weight values}
\begin{figure}[h!]
      \centering
      \includegraphics[scale=0.36, trim={0, 0, 0, 0}, clip]{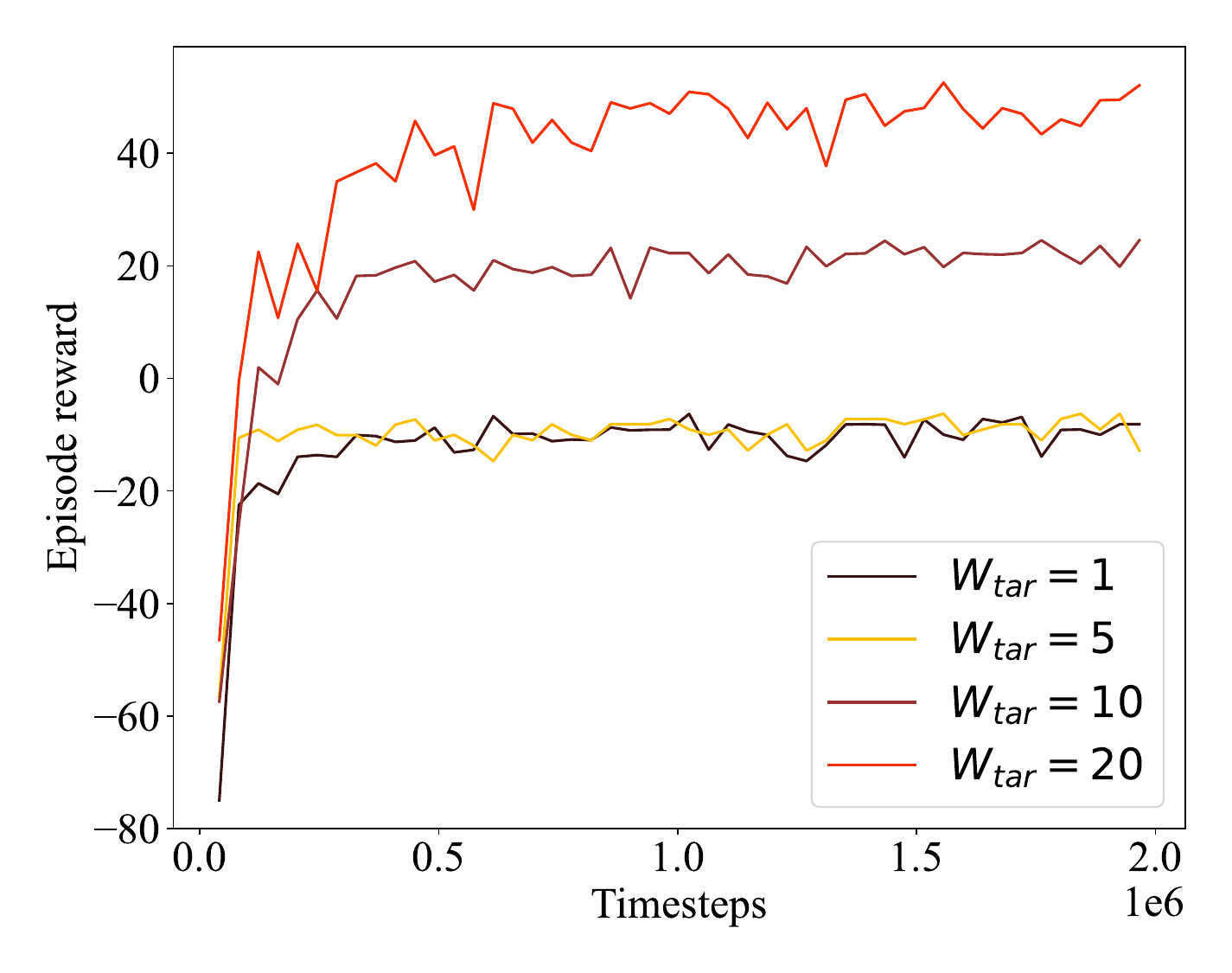}
      \caption{Learning of TCOP-based rewards in the new architecture with PPO using different $W_{tar}$ values and $W_c$, $W_{nc}$, $W_o$ = 0.1.}
      \label{fig:tcop_rew}
   \end{figure}

In this analysis, we apply the TCOP-based reward function in \autoref{eq:tcop_rew} to the new architecture in order to assess whether the agent can successfully acquire a driving strategy that is both safe and cost-effective. The episodic reward is evaluated with varying weights $W_{tar}$ assigned to the target completion reward $R_{tar}$, as depicted in Fig. \ref{fig:tcop_rew}. As mentioned in Section II, in an ideal scenario, the cost would amount to 2.315 euros, while the revenue ($R_{tar}$) would be 2.78 euros. Consequently, in this ideal case, the episodic reward would equal $2.78 - 2.315 = 0.46$ when $W_{tar}$ is 1. This value is relatively small, making it challenging for the agent to learn how to reach the target. Furthermore, the penalties for hazardous situations such as collision, near collision and driving outside the road are 1000 euros which is comparatively very high. This leads to a risk for the agent to focus only on avoiding such situations and do not care about reaching the target, for example by driving with very low speed. This issue can be overcome by carefully choosing the weight values for the penalty and revenue components. We observed good results by keeping the weight value for penalties $W_c$, $W_{nc}$, $W_o$ = 0.1 and increasing the weight value for revenue $W_{tar}$ to 20 as illustrated in \autoref{fig:tcop_rew}. From \autoref{tab:tcop_rew}, it becomes evident that increasing the weight $W_{tar}$ assists the agent in learning to drive at higher speeds and successfully reach the target. Notably, as the weight $W_{tar}$ is increased, the total cost of operation—comprising energy cost and driver cost—is minimized to 3.74 euros. This cost is lower than the cost incurred when the ego vehicle was entirely controlled by SUMO, using the default Krauss car following model and LC2013 lane change model in the same traffic environment, which resulted in a cost of 3.85 euros.
\begin{table}[h!]
\caption{Evaluation of new architecture with TCOP based reward function, using different $W_{tar}$ values and $W_c$, $W_{nc}$, $W_o$ = 0.1}
\label{tab:tcop_rew}
\begin{tabular}{|p{3cm}|p{1.3cm}|p{1.3cm}|p{1.3cm}|p{1.3cm}|}
\hline
\textbf{Evaluation Metric} & \boldmath{$W_{tar} = 1$} & \boldmath{$W_{tar} = 5$}  & \boldmath{$W_{tar} = 10$} & \boldmath{$W_{tar} = 20$}\\
\hline
Reached target successfully & 2.2\% & 13.6\% & 68.2\% & 99.2\%
\\ 
\hline
Driven successfully, but not reached the target within maximum steps & 95.2\% & 83.6\% & 29.6\% & 0\%
\\ 
\hline
Terminated by collision or driving outside road & 2.6\% & 2.8\% & 2.2\% & 0.8\%
\\ 
\hline
Average speed & 1.75 m/s & 2.78 m/s & 11.16 m/s & 18.36 m/s
\\ 
\hline
Average distance travelled & 615.9 m & 848.2 m & 1864.7 m & 2197.4 m
\\ 
\hline
Average executed steps & 482 & 455 & 265 & 122
\\ 
\hline
Average energy cost & 0.17 euros & 0.28 euros & 1.1 euros & 2.05 euros
\\ 
\hline
Average driver cost & 6.7 euros & 6.32 euros & 3.68 euros & 1.69 euros
\\ 
\hline
Average tcop & 6.87 euros & 6.6 euros & 4.78 euros & 3.74 euros
\\ 
\hline
\end{tabular}
\end{table}

\subsubsection{Comparison with and without reward normalization in CRL and non-CRL}
In these experiments, instead of adding weights to the reward components, we investigate whether the agent can learn an optimal policy with the modified TCOP reward function as in \autoref{eq:tcop_rew_new}
that contains only the actual costs and revenue values. We compare the performance with and without CRL approach using PPO algorithm. PPO was chosen because it proved to give consistent performance in earlier experiments. To improve exploration, we used an adaptive entropy coefficient of 0.01 which gradually decreases to 0.001 whereas default values in stable-baselines3 are used for all other hyperparameters.

\begin{figure}[h!]
\centering
\includegraphics[scale=0.47, trim={0, 0, 0, 2mm}, clip]{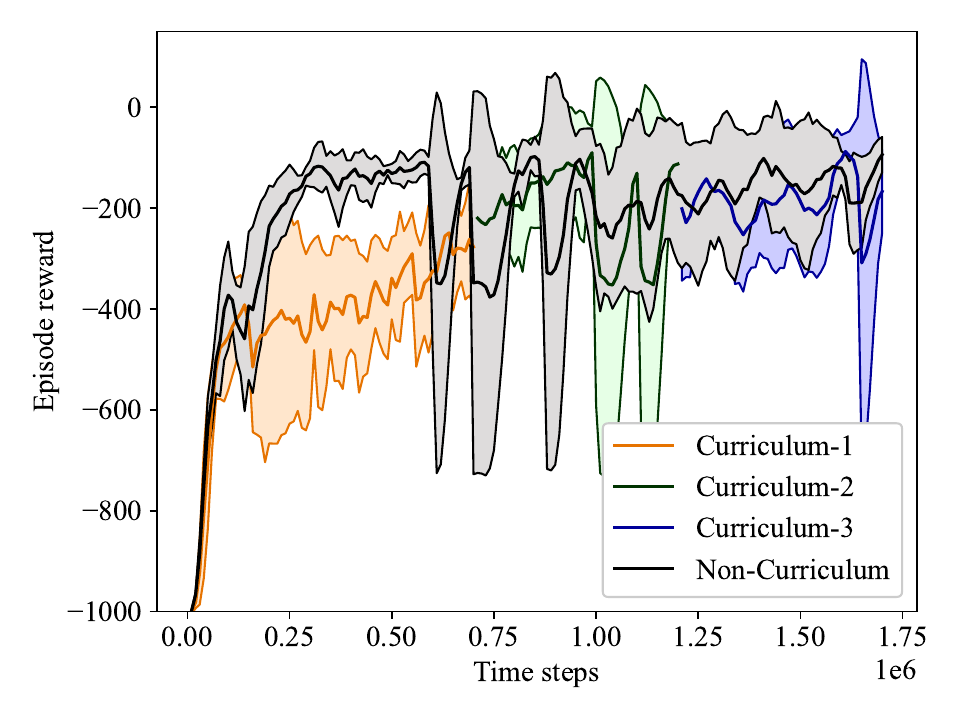}
 \caption{Learning of TCOP based reward function without normalization using CRL and non-CRL approaches.}
 \label{fig:curr}
\end{figure}

\begin{table}[h!]
\caption{Evaluation of agent trained with TCOP based reward function without normalization using CRL and non-CRL approaches.}
\label{tab:curr}
\begin{tabular}{|p{3.4cm}|p{1.9cm}|p{1.9cm}|p{1.9cm}|p{1.9cm}|}
\hline
\textbf{Evaluation Metric} & \textbf{Curriculum-1} & \textbf{Curriculum-2} & \textbf{Curriculum-3} & \textbf{Non-Curriculum}
\\ \hline Reached target successfully & 44.80\% & 10.35\% & 7.15\% & 2.50\%
\\ \hline Terminated by collision or driving outside road & 13.45\% & 8.80\% & 9.60\% & 6.20\%
\\ \hline Driven successfully, but not reached the target within maximum steps & 41.75\% & 80.85\% & 83.25\% & 91.30\%
\\ \hline Average distance travelled & 1269.80 m & 837.52 m & 739.12 m & 603.66 m
\\ \hline Average executed steps & 265.91 & 431.43 & 438.64 & 463.48
\\ \hline Average speed & 11.38 m/s & 3.80 m/s & 2.99 m/s & 2.14 m/s
\\ \hline Average energy cost & 1.23 euros & 0.38 euros & 0.27 euros & 0.20 euros
\\ \hline Average driver cost & 3.7 euros & 6 euros & 6.1 euros & 6.44 euros
\\ \hline Average tcop & 4.93 euros & 6.38 euros & 6.37 euros & 6.64 euros
\\ \hline
\end{tabular}
\end{table}

\begin{figure}[h!]
\centering
\includegraphics[scale=0.47, trim={0, 0, 0, 2mm}, clip]{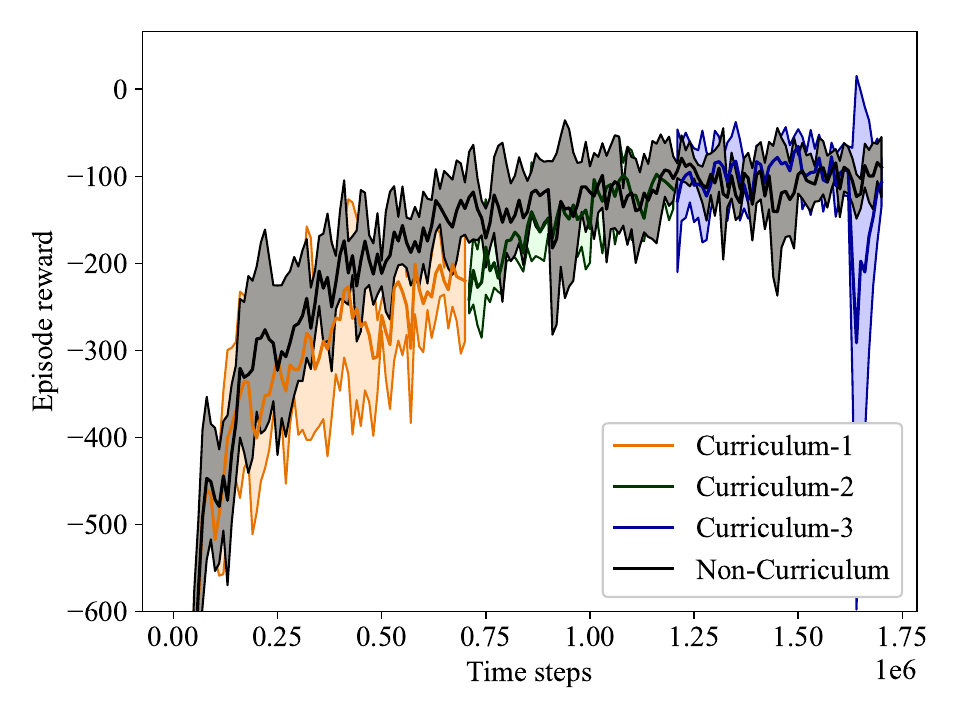}
 \caption{Learning of TCOP based reward function with normalization using CRL and non-CRL approaches.}
 \label{fig:curr_normalized}
\end{figure}

\begin{table}[h!]
\caption{Evaluation of agent trained with TCOP based reward function with normalization using CRL and non-CRL approaches.}
\label{tab:curr_normalized}
\begin{tabular}{|p{3.4cm}|p{1.9cm}|p{1.9cm}|p{1.9cm}|p{1.9cm}|}
\hline
\textbf{Evaluation Metric} & \textbf{Curriculum-1} & \textbf{Curriculum-2} & \textbf{Curriculum-3} & \textbf{Non-Curriculum}
\\ \hline Reached target successfully & 57.45\% & 49.15\% & 73.50\% & 73.25\%
\\ \hline Terminated by collision or driving outside road & 8.25\% & 4.45\% & 2.90\% & 3.50\%
\\ \hline Driven successfully, but not reached the target within maximum steps & 34.30\% & 46.40\% & 23.60\% & 23.25\%
\\ \hline Average distance travelled & 1569.99 m & 1382.20 m & 1794.74 m & 1770.03 m
\\ \hline Average executed steps & 252.24 & 294.00 & 210.22 & 206.94
\\ \hline Average speed & 12.27 m/s & 10.38 m/s & 14.18 m/s & 14.42 m/s
\\ \hline Average energy cost & 1.33 euros & 1.08 euros & 1.68 euros & 1.62 euros
\\ \hline Average driver cost & 3.50 euros & 4.08 euros & 2.92 euros & 2.88 euros
\\ \hline Average tcop & 4.83 euros & 5.17 euros & 4.60 euros & 4.50 euros
\\ \hline
\end{tabular}
\end{table}

\begin{figure}[h!]
\begin{subfigure}{0.49\textwidth}
     \includegraphics[width=\textwidth, trim={0, 0, 0, 0}, clip]{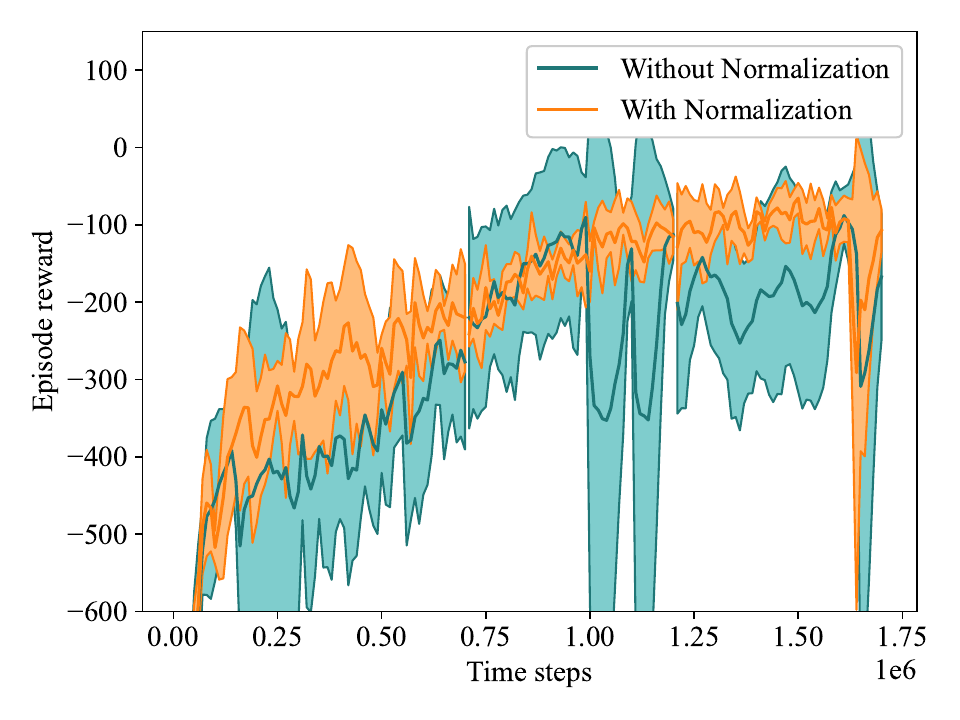}
     \caption{CRL}
     \label{fig:base_arch_res}
 \end{subfigure}
 \hfill
 \begin{subfigure}{0.49\textwidth}
     \includegraphics[width=\textwidth]{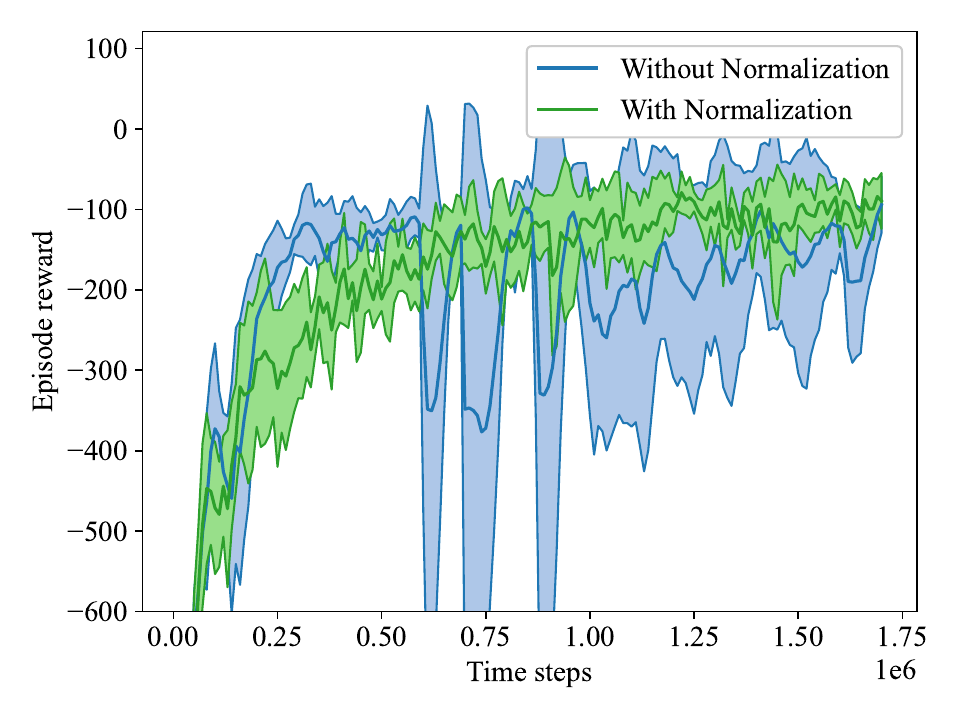}
     \caption{Non-CRL}
     \label{fig:new_arch_res}
 \end{subfigure}
 \caption{Performance comparison with and without normalization in TCOP based rewards}
 \label{fig:norm_comp}
\end{figure}

\begin{table}[h!]
\caption{Comparison of validation results with and without normalization in TCOP based rewards in CRL and non-CRL approaches.}
\label{tab:comp_rew}
\begin{tabular}{|p{3.4cm}|p{1.9cm}|p{1.9cm}|p{1.9cm}|p{1.9cm}|}
\hline
 & \multicolumn{2}{c|}{\textbf{CRL}} & \multicolumn{2}{c|}{\textbf{Non-CRL}} \\ \hline
\textbf{Evaluation Metric} & \textbf{Without normalization} & \textbf{With Normalization} & \textbf{Without normalization} & \textbf{With Normalization} \\ \hline
Reached target successfully & 7.15\% & 73.50\% & 2.50\% & 73.25\% \\ \hline
Terminated by collision or driving outside road & 9.60\% & 2.90\% & 6.20\% & 3.50\% \\ \hline
Driven successfully, but not reached the target within maximum steps & 83.25\% & 23.60\% & 91.30\% & 23.25\% \\ \hline
Average distance travelled & 739.12 m & 1794.74 m & 603.66 m & 1770.03 m \\ \hline
Average executed steps & 438.64 & 210.22 & 463.48 & 206.94 \\ \hline
Average speed & 2.99 m/s & 14.18 m/s & 2.14 m/s & 14.42 m/s \\ \hline
Average energy cost per meter & 0.00036 euros & 0.00094 euros & 0.00033 euros & 0.00092 euros \\ \hline
Average driver cost per meter & 0.0082 euros & 0.0016 euros & 0.0107 euros & 0.0016 euros \\ \hline
Average tcop per meter & 0.0086 euros & 0.0026 euros & 0.011 euros & 0.0025 euros \\ \hline
\end{tabular}
\end{table}

\autoref{fig:curr} shows the episodic reward training curve for the RL agent trained with curriculum learning and compares it to the training curve without using curriculum learning. Here, the agent is trained with curriculum learning using the reward functions without normalization given in \autoref{eq:cl_rew} where curriculum-1 is trained for $7e^5$ time steps and curriculum-2 and curriculum-3 are trained for $5e^5$ time steps each. Non-curriculum learning results are obtained from RL agent trained with the reward function \autoref{eq:tcop_rew_new} for $17e^5$ time steps. There are noticeable drops in the reward curve at certain points during training in both curriculum and non-curriculum learning. The validation results in \autoref{tab:curr} also show a poor success rate. Here, the evaluation is performed by averaging the validation results from 5 trained models for curriculum and non-curriculum approaches.

It can be seen that the average speed of the ego vehicle drops down largely after training with curriculum-2 when the penalty for energy cost is introduced, which also leads to a reduction in the success rate. This motivates us to tune the reward function by normalizing driver cost and energy cost by distance travelled per time step. \autoref{fig:curr_normalized} and \autoref{tab:curr_normalized} show the results of the agent trained with this normalized reward function in \autoref{eq:cl_rew_norm} and corresponding non-curriculum learning. Here, the reduction in speed from curriculum-1 to curriculum-2 is smaller and this reduction is recovered when the reward for target completion is added to the next curriculum. Consequently, the success rate is higher in last curriculum compared to the previous case. \autoref{fig:norm_comp} compares the training curves with and without reward normalization where we could observe that the average reward is comparatively higher with normalization in both CRL and non-CRL approaches. The training curve with normalization also shows a more stable and steady improvement in rewards, even though the drop in the last curriculum persists. Overall, the normalized reward function significantly improves the results for both CRL and non-CRL approaches as shown explicitly in \autoref{tab:comp_rew}. We could observe significant improvement in success rate, average speed and TCOP. However, the results obtained for curriculum and non-curriculum approaches are very similar which demonstrates that there is no particular advantage in using curriculum learning in this RL setting.


\section{Conclusion}
We implemented an RL framework for tactical decision decision making of autonomous trucks in a highway environment by integrating RL with low-level controllers. Our results demonstrate that training the agent to focus on high level decisions, such as maintaining a time gap with the leading vehicle, while leaving the low level speed control to a physics-based controller, accelerates learning and improves overall performance. Additionally, we explored the effectiveness of training the agent with realistic rewards and penalties using a multi-objective TCOP based reward function. We study this setting with different approaches, by adding weights to reward components, by normalizing the reward components and by using curriculum learning techniques. We could observe that reshaping the reward function with weights or normalization significantly improves the performance whereas CRL shows comparable results with non-CRL approach.

An interesting future direction would be to explore transfer learning methods for generalizing tactical decision-making to diverse traffic scenarios such as uphill, downhill, merging traffic etc. In this context, we believe that the pre-trained models from simple highway environments can serve as a strong foundation while adapting to new environments with different dynamics and constraints.

\section*{Acknowledgment}
This work was partially supported by Wallenberg AI, Autonomous Systems and Software Program (WASP) funded by Knut and Alice Wallenberg Foundation. We would like to thank Nikolce Murgovski, Erik Börve and Stefan Börjesson for valuable discussions.

\section*{Declarations}
\paragraph{Confict of interest} There is no confict of interest.
\paragraph{Ethics approval} Not applicable.
\paragraph{Consent to participate} Not applicable.
\paragraph{Consent for publication} Not applicable.
\paragraph{Code availability} The source code is available at: \\ \url{https://github.com/deepthi-pathare/Autonomous-truck-sumo-gym-env}

\bibliography{sn-bibliography}

\end{document}